\documentclass[11pt,letterpaper]{article}
\usepackage[margin=1in]{geometry}
\usepackage[T1]{fontenc}
\usepackage{lmodern}
\usepackage{amsmath,amssymb}
\usepackage{graphicx,booktabs,tabularx,adjustbox}
\usepackage[numbers,sort&compress]{natbib}

\usepackage{microtype}
\usepackage[section]{placeins}
\usepackage[colorlinks=true,linkcolor=blue,citecolor=blue,urlcolor=blue]{hyperref}
\providecommand{\Description}[1]{}
\ifdefined\pdfinfoomitdate \pdfinfoomitdate=1 \fi
\ifdefined\pdfsuppressptexinfo \pdfsuppressptexinfo=-1 \fi
\ifdefined\pdftrailerid \pdftrailerid{} \fi
\title{BAIBAICHUCHU at the NTCIR-19 FinArg-3 Task: When Is Maximum Possible Profit Predictable from Investor Text?}
\author{Zong-Han Bai\thanks{Both authors contributed equally to this research.}\\
National Tsing Hua University\\Hsinchu, Taiwan\\
\texttt{hu110048138@gapp.nthu.edu.tw}
\and Po-Yen Chu\footnotemark[1]\\
Stanford University\\Stanford, CA, USA\\
\texttt{poyenchu@stanford.edu}}
\date{}
\hypersetup{pdftitle={BAIBAICHUCHU at the NTCIR-19 FinArg-3 Task: When Is Maximum Possible Profit Predictable from Investor Text?},pdfauthor={Zong-Han Bai and Po-Yen Chu}}

\begin{document}
\maketitle
\begin{abstract}
The BAIBAICHUCHU team participated in the Social Media Subtask of the NTCIR-19 FinArg-3 Task, which asks which of two Chinese investor posts attains the higher Maximum Possible Profit (MPP). Our three-track ensemble---lexical features, a FinArg-2-pre-finetuned MacBERT ranker, and an LLM judge with a market-semantics rubric---reaches 0.734 accuracy in post-grouped development evaluation. Yet our best official run falls to 0.517, all eleven NTCIR-19 submitted runs lie between 0.4598 and 0.5402, and even the 26 official pairs on which our three tracks agree unanimously score exactly 0.500. We treat this development-to-test collapse as the paper's central puzzle.

A post-hoc audit reveals that our submitted judge treated bearish posts under a long-only payoff rule although MPP is defined stance-aware. Correcting the rule changes 28 of 87 official predictions without changing accuracy, yet lowers in-domain accuracy from 0.680 to 0.622: a regime-specific shortcut improved validation fit while violating the intended semantics.

We decompose profitability ranking into directional text, an outcome scale set by volatility and horizon, and pairwise margin. A minimal running-extremum model predicts $\sigma\sqrt{T}$ scaling, observed ex post among 502 price-aligned posts from a separate 2{,}100-post July 2026 collection. Using pre-posting volatility and the post-specific July observation horizon, $\sigma_{\mathrm{pre},20}\sqrt{N_i}$ is associated with a later truncated-horizon MPP proxy (TH-MPP; Spearman $\rho=0.320$, ticker-cluster 95\% CI $[0.133,0.466]$) and correctly orders 61.4\% of the unequal-outcome July post pairs. The transferred text score is nearly uncorrelated with the July outcome ($\rho=0.055$) and adds little conditional on this score and stance. Separately, a one-parameter reliability model shows development accuracy rising from about 0.60 to 0.93 as the labeled MPP gap widens. Since FinNLP-2022 ERAI reported 0.6207 MPP-ranking accuracy, the NTCIR-19 collapse does not imply universal unpredictability; these results motivate jointly considering text, market regime, pre-posting volatility, observation horizon, and pair composition.
\end{abstract}


\par\medskip\noindent\textbf{Team Name:} BAIBAICHUCHU\\
\textbf{Subtasks:} Social Media Subtask (Chinese)

\section{Introduction}

The Social Media Subtask of NTCIR-19 FinArg-3~\cite{chen2026overview} presents systems with pairs of Chinese social-media posts about Taiwanese stocks and asks which post attains a higher Maximum Possible Profit (MPP), a market-outcome label introduced by Chen et al.~\cite{chen2021rationales}. Although the task is motivated by investment argument quality, the label is realized through a future price path. Profitability ranking therefore sits at an unusual boundary between language understanding and market-outcome prediction.

Our results expose a sharp transfer puzzle. A heterogeneous three-track ensemble reaches 0.734 accuracy in post-grouped development evaluation, but our best official run reaches only 0.517. The effect is not specific to our submission: all eleven NTCIR-19 runs fall between 0.4598 and 0.5402, and the 26 official pairs on which our three tracks agree unanimously achieve exactly 0.500 accuracy. At the same time, the historical task family is not uniformly impossible: the FinNLP-2022 ERAI predecessor reported a 0.6207 MPP-ranking result with BERT-Chinese~\cite{gon2022jetsons}. The contrast rules out a simple ``MPP cannot be predicted from text'' explanation and instead motivates a study of why apparent signal fails to transfer across evaluation regimes.

A first diagnosis came from an error in our own submission. Prior work defines MPP explicitly as stance-aware~\cite{chen2021rationales}, but our submitted LLM judge incorrectly used a long-only treatment of bearish posts. Correcting the semantics changes many predictions yet does not improve official accuracy. In development, however, the correction actually reduces judge accuracy from 0.680 to 0.622. The error therefore becomes informative: the wrong rule encoded a useful prior for a predominantly bullish training regime. This shows \emph{semantic fidelity and empirical validation fit diverging}, a failure mode particularly relevant when LLM judges are tuned through iterative error analysis.

We then ask what else determines whether a pair is rankable. \emph{Directional text signal} concerns what the author says. \emph{Pre-posting volatility and observation horizon} set the model-implied excursion scale. \emph{Pairwise margin} is the distance between realized outcomes: for a fixed noisy predictor, narrow-margin pairs are harder to order. These components structure what distribution shift can change without replacing it as an explanation of the official collapse.

The paper makes four contributions. \emph{First, system and transfer puzzle}: a heterogeneous ensemble reaches 0.734 in post-grouped development yet collapses to near chance on the official set, motivating an analysis of transfer rather than leaderboard optimization alone. \emph{Second, semantic shortcut}: correcting the bearish payoff definition lowers development judge accuracy while leaving official accuracy unchanged, showing that semantic fidelity and regime-specific predictive fit can diverge. \emph{Third, ex-ante volatility}: we derive the $\sigma\sqrt{T}$ scaling and show that $\sigma_{\mathrm{pre},20}\sqrt{N_i}$ is associated with the later July proxy outcome and pairwise ordering. \emph{Fourth, benchmark decomposition}: transferred textual information is weak after controlling for this pre-posting volatility--horizon score, while pairwise reliability depends on the realized outcome margin, so observed performance reflects model signal, outcome structure, and pair construction.\footnote{Code and analysis outputs: \url{https://github.com/HummerQAQ/finarg3-social-media}.}
\section{Related Work}
FinArg has progressively broadened financial argument mining from argument-based sentiment and relation analysis in FinArg-1~\cite{lin2024argument}, through temporal inference in FinArg-2~\cite{finarg2overview}, to argument-quality assessment in FinArg-3. For social media, FinArg-3 asks which of two investor posts attains the higher MPP, following the profitability-based evaluation introduced by Chen et al.~\cite{chen2021rationales} and later extended to investment-opinion ranking~\cite{chen2025opinion}.

A direct historical predecessor is the FinNLP-2022 ERAI shared task~\cite{chen2022erai}, which likewise evaluated Chinese investor posts with MPP/ML labels and included pairwise MPP comparison. The Jetsons BERT-Chinese system reported 0.6207 accuracy on the ERAI blind MPP test~\cite{gon2022jetsons}. This historical contrast is central to our interpretation: the task family has previously admitted substantially above-chance ranking, whereas every NTCIR-19 submitted run lies close to chance. Public materials do not establish pair-level identity between ERAI and FinArg-3, so we treat the contrast as motivation rather than as a controlled comparison.

Two organizer publications motivate our modeling choices. Pre-finetuning on professionalism labels improves profitability ranking on this data family~\cite{chen2024professionalism}, while duration-aware pre-finetuning transfers to stock-movement prediction~\cite{chiu2025prefinetuning}. Our encoder track follows the latter strategy, using the FinArg-2 validity-duration corpus as a same-domain auxiliary task. At NTCIR-18, fine-tuned Chinese PLMs and a fine-tuned GPT-4o-mini were separated by small margins on the social-media subtask~\cite{imntpu2025}; our backbone is MacBERT~\cite{cui2020macbert}, a Chinese BERT-family model~\cite{devlin2019bert}.

Our judge track also inherits known pairwise-LLM failure modes. Zheng et al.~\cite{zheng2023judging} document presentation-order and verbosity biases; we query both orders, use order agreement as a confidence signal, and allow abstention on disagreement. Finally, Section~\ref{sec:theory} uses the efficient-market perspective~\cite{fama1970efficient} not as an assumption of zero predictability but to distinguish directional textual information from publicly observable market covariates, state revelation, and possible non-public information.
\section{What Does the Label Measure?}
\label{sec:semantics}
The training release contains 210 rows. After removing 57 rows with missing labels and deduplicating identical texts (duplicates always agree on MPP), 91 unique labeled posts remain. The official test set contains 87 pairs over 174 distinct posts with no overlap with training. The two distributions differ sharply in register: training posts are long forum theses (median 210 characters), test posts are short casual comments (median about 70 characters, 43\% under 50).

\FloatBarrier
\subsection{Stance handling: a submitted-system error audited with external prices}
\label{sec:stance}

Prior work defines MPP explicitly as stance-aware~\cite{chen2021rationales}: favorable excursion is upward for bullish opinions and downward for bearish opinions. The literature is therefore not ambiguous on this point. The released scalar MPP/ML values, however, are not self-describing with respect to that mapping when viewed without the underlying price path. During development, we attempted to infer the mapping from the released values and incorrectly implemented a long-only treatment of bearish posts in the LLM judge.

External prices\footnote{Daily prices from Yahoo Finance via \texttt{yfinance} (Taiwanese listings as \texttt{<code>.TW}, fallback \texttt{.TWO}); dividend-unadjusted highs, lows, and closes. The training release has no posting timestamps, so the audit aligns dates from author-quoted prices and keeps only cases where the competing interpretations imply clearly incompatible price paths. Accessed August 2026.} confirm the published mapping. One bearish Macronix post records $\mathrm{ML}=-0.369$ from an entry price of 23.8; over the corresponding window the stock later traded as high as 32.85, consistent with the adverse excursion of a short position. The released labels are therefore consistent with
\[
\mathrm{MPP}_{\mathrm{bull}}=\frac{\max H-O}{O},
\qquad
\mathrm{MPP}_{\mathrm{bear}}=\frac{O-\min L}{O}.
\]
Section~\ref{sec:stancefix} quantifies the effect of correcting our judge.

Three observations survive the correction: bearish posts receive low MPP in this training sample, more naturally a property of the market regime than of the label; index-related vocabulary correlates with MPP at $r=-0.27$, a descriptive asset-class association whose mechanism is not identified here; and text can reveal contemporaneous market state through recent moves, drawdowns, and entry context.

\FloatBarrier
\subsection{Textual correlates}

To characterize what information in the posts is associated with MPP, we used an LLM to extract structured attributes from each of the 91 labeled posts, including target asset class, short-term stance, momentum and state cues, catalyst timing, chip-flow signals, insider hints, volatility tier, narrative type, and causal-chain completeness. We then fit a pairwise logistic regression over differences in these attributes. Table~\ref{tab:attrs} reports a selected set of standardized coefficients; given the small sample and correlated features, they should be read as directional evidence rather than causal effects.

\begin{table}[htbp]
\centering
\small
\caption{Selected coefficients of LLM-extracted attributes under the pairwise objective. Features are standardized and correlated; coefficients are exploratory directional evidence at $n=91$, not causal estimates.}
\label{tab:attrs}
\begin{adjustbox}{max width=\linewidth}
\begin{tabular}{lr}
\toprule
Attribute & Weight \\
\midrule
Targets a single stock (vs.\ index / no asset) & $+0.88$ \\
Insider hint (``dream post'' idiom) & $+0.54$ \\
Short-term bullish stance & $+0.43$ \\
LLM upside score (legacy framing) & $+0.34$ \\
Narrative: growth & $+0.30$ \\
Known catalyst within days & $-0.32$ \\
Narrative: turnaround & $+0.02$ \\
Causal-chain completeness & $+0.00$ \\
\bottomrule
\end{tabular}
\end{adjustbox}
\end{table}

The single-stock indicator has the strongest coefficient, but development data do not establish that volatility mediates this descriptive asset-class association. Insider hints also receive a large positive weight ($r=+0.40$ in isolation), suggesting that some textual signal may arise from information unavailable in conventional argument-quality features. The positive coefficient on bullish stance is better interpreted as a property of the predominantly bullish training regime than of the MPP definition itself.

By contrast, conventional argument-quality features contribute little. Turnaround narratives and causal-chain completeness receive essentially zero weight, while proximity to a publicly known catalyst is negatively associated with MPP. These patterns are exploratory at this sample size, but they motivate the later decomposition: predictive signal may come from target-instrument identity, market state, and information content rather than rhetorical completeness alone.

\FloatBarrier
\subsection{Illustrative cases}
\label{sec:cases}
The training data illustrate why rhetorical quality alone is insufficient for predicting MPP. A short ``dream post'' hinting at possibly non-public information attains substantially higher MPP than a long valuation analysis; a detailed bearish thesis loses to a simpler bullish contrarian post when the downside excursion never materializes; and a post written during a sharp drawdown reveals contemporaneous market state even without sophisticated causal reasoning.

\section{System}
\label{sec:system}
Our system combines three tracks with deliberately different inductive biases: lexical market cues, learned semantic representations, and explicit LLM-based market reasoning. Their predictions are combined by majority vote. Because design choices were iteratively informed by development errors, all cross-validation scores below are tuned development estimates rather than unbiased generalization estimates.

\FloatBarrier
\subsection{Evaluation protocol}
Because each post appears in many pairs, random pair-level splitting would leak post-level information across train and validation sets. We assign the 91 unique posts to five folds stratified by MPP rank, construct all cross-post pairs with $|\Delta\mathrm{MPP}|\ge0.02$ (3{,}591 pairs), and evaluate fold $k$ only on pairs whose two posts both belong to fold $k$, while training only on pairs whose posts lie outside it. Repeating the fold assignment under three additional seeds changes accuracy by about $\pm0.03$, so gains below two points were treated as development noise. The linear track additionally weights samples by $|\Delta\mathrm{MPP}|$, worth $+1.8$ points in ablation. The five LLM few-shot exemplars were chosen during error analysis; pairs containing them are excluded from the 553-pair common ensemble subset, though rubric and lexicon design remain development-tuned.

\FloatBarrier
\subsection{Track 1: Lexicon features and character n-grams}

The first track is a logistic regression over character 1--3-gram TF-IDF concatenated with roughly fifty handcrafted features. These include Taiwanese stock-forum lexicons for fundamentals, technical analysis, chip flows, bullish and bearish stance, hedging, emotion, index and macro vocabulary, and low-volatility instruments, together with numeral, price, and URL statistics. A pair $(a,b)$ is represented antisymmetrically as $f(a)-f(b)$. The index-macro lexicon, added after manual inspection of the training posts, is the single largest contributor, improving accuracy by $3.5$ points.

\FloatBarrier
\subsection{Track 2: Pre-finetuned MacBERT ranker}

The second track learns a scalar score $s(x)$ for each post and minimizes $\mathrm{BCE}(s(a)-s(b),y)$ over training pairs. This formulation is antisymmetric by construction, avoids repeated cross-encoding of the same post, and yields reusable per-post scores. Following the organizers' pre-finetuning recipe~\cite{chiu2025prefinetuning}, we first train MacBERT on validity-duration classification over 8{,}760 FinArg-2 social media posts~\cite{finarg2overview}, then fine-tune it for pairwise MPP ranking. The auxiliary corpus has a median length of 54 characters, which also adapts the encoder toward the shorter test-post register. Five random seeds are ensembled by averaging z-normalized scores.

\FloatBarrier
\subsection{Track 3: LLM judge with a market-semantics rubric}

The third track uses GPT-5-mini as a pairwise judge under a Traditional Chinese rubric derived from the empirical patterns in Section~\ref{sec:semantics}. The rubric asks the model to identify the target asset, attend to momentum and market-state cues, discount purely post-hoc explanations and widely known catalysts, and recognize compressed technical summaries and possible insider-hint idioms. The submitted rubric also encoded our initial, incorrect interpretation of bearish payoff handling; a stance-corrected variant is evaluated separately in Section~\ref{sec:stancefix}. Five few-shot exemplars are selected from real training pairs identified during earlier error analysis.

Each pair is queried in both presentation orders, and the resulting probabilities are aligned and averaged. Order agreement also provides a confidence signal: validation accuracy is 0.71 when the two orderings agree but only 0.53 when they disagree. We therefore let the judge abstain on order-inconsistent pairs and fall back to the mean prediction of the two local tracks.

\FloatBarrier
\subsection{Runs}
Run 1 is the consistency-filtered majority vote. Run 2 is a blend of the two local tracks only, immune to API failure. Run 3 replaces the judge with a learned model over LLM-extracted attributes, providing a judge-free diversity variant.

\section{Development Results and Negative Ablations}
\label{sec:ablations}

\FloatBarrier
\subsection{How the rubric was built}
\label{sec:rubric}
Table~\ref{tab:rubric} summarizes the judge-rubric iterations.

\begin{table}[htbp]
\centering
\small
\caption{Rubric iterations on the shared judge validation pairs.}
\label{tab:rubric}
\centering
\begin{tabularx}{\columnwidth}{lXc}
\toprule
Version & Change & Acc. \\
\midrule
v1   & argument-quality rubric & 0.520 \\
v2   & market-outcome reframing + counterintuitive few-shots & 0.599 \\
v3   & + eight error-derived rules & \textbf{0.680} \\
v3.1 & + qualifiers against over-correction & 0.666 \\
\bottomrule
\end{tabularx}
\end{table}

The initial argument-quality rubric favored longer posts and scored 0.520. Reframing the task around the market outcome and adding counterintuitive few-shots raised accuracy to 0.599. We then reviewed twenty confident errors and encoded eight recurring failure modes, from target-asset confusion to bearish payoff handling; rubric v3 reached 0.680. Additional qualifiers reduced accuracy to 0.666, so v3 was retained.

\FloatBarrier
\subsection{Main results}

\begin{table}[htbp]
\centering
\small
\caption{Post-grouped development accuracy. The lexical and encoder tracks use 717 evaluation pairs; the LLM judge and ensemble rows use a 553-pair common subset that excludes pairs containing few-shot exemplar posts.}
\label{tab:cv}
\begin{adjustbox}{max width=\linewidth}
\begin{tabular}{lc}
\toprule
System & Dev. accuracy \\
\midrule
Feature baseline (pre-lexicon update) & 0.647 \\
Track 1: lexicon + n-gram LR & 0.682 \\
Track 2: MacBERT, no pre-finetuning (6 seeds) & 0.668 \\
Track 2: + duration pre-finetuning (5 seeds) & 0.682 \\
Track 3: LLM judge, rubric v3 & 0.680 \\
\midrule
Majority vote (three tracks) & 0.731 \\
+ order-consistency filter (Run 1) & \textbf{0.734} \\
\bottomrule
\end{tabular}
\end{adjustbox}
\end{table}

Table~\ref{tab:cv} summarizes the adopted system. Duration pre-finetuning changes the MacBERT ensemble from 0.668 to 0.682 overall and from 0.683 to 0.697 on short-post pairs, but the overall $+1.4$-point gain is below our two-point development-noise threshold and should be treated as directional rather than conclusive. The three tracks exhibit only moderate pairwise correlation (0.32--0.55), motivating the majority-vote design; on the common 553-pair subset, the majority reaches 0.731. The order-consistency filter adds a smaller gain, improving accuracy from 0.649 to 0.670 on the subset of pairs for which the judge's two presentation orders disagree.

\FloatBarrier
\subsection{Negative ablations}

\begin{table}[htbp]
\centering
\small
\caption{Negative ablations. ``ns'' denotes changes within the $\pm0.02$ development-noise range.}
\label{tab:negative}
\centering
\small
\begin{tabularx}{\columnwidth}{Xcc}
\toprule
Variant & Result & Relative effect \\
\midrule

\multicolumn{3}{l}{\emph{Judge variants (adopted: rubric v3, GPT-5-mini, 0.680)}} \\
Quality-prior rubric (v1) & 0.520 & $-0.160$ \\
Over-correction qualifiers (v3.1) & 0.666 & $-0.014$ \\
Stance-corrected rubric (post-hoc) & 0.622 & $-0.058$ \\
GPT-5.1, default reasoning & 0.658 & $-0.022$ \\
GPT-5.1, high reasoning & 0.680 & $\pm0$ \\
kNN-retrieved exemplars & 0.680 & $\pm0$ \\

\midrule
\multicolumn{3}{l}{\emph{Encoder variants (adopted: duration pre-finetuning, 0.682)}} \\
+ argument-relation second stage & 0.657 & $-0.025$ \\
Margin-weighted loss & 0.671 & $-0.011$ \\
Soft pair targets ($\tau=0.05$) & 0.661 & $-0.021$ \\
Degree-normalized pairs & 0.683 & ns \\

\midrule
\multicolumn{3}{l}{\emph{Alternative supervision, linear track (adopted: 0.682)}} \\
Soft pair targets & 0.640 & $-0.042$ \\
Pointwise MPP regression & 0.656 & $-0.026$ \\
Pairwise gap regression & 0.646 & $-0.036$ \\

\midrule
\multicolumn{3}{l}{\emph{Ensemble and aggregation (adopted majority: 0.734)}} \\
Frozen e5/bge embedding rankers & 0.646/0.639 & diluting \\
LLM-attribute model as 4th voter & 0.638 & diluting \\
Weighted probability fusion & 0.69--0.71 & $\approx-0.02$ \\
Selective arbitration cascade & 0.723 & $-0.011$ \\
Human overrides & fixes 4 / breaks 8 & net negative \\

\bottomrule
\end{tabularx}
\end{table}

Table~\ref{tab:negative} summarizes the main negative ablations; three patterns recur. Judge accuracy plateaus near 0.68: further qualifiers, retrieved exemplars, and larger or more deliberate models do not improve it consistently (the stance-corrected variant is analyzed in Section~\ref{sec:stancefix}). The form of supervision matters more than its amount: using $|\Delta\mathrm{MPP}|$ as a sample weight helps the linear model, whereas soft targets and direct regression on MPP or gaps hurt, so margin information is better used to modulate directional supervision than to replace the ranking target. Ensemble gains come from complementary errors: frozen embedding rankers correlate 0.80--0.88 with the MacBERT track and dilute the vote, whereas the judge contributes despite similar standalone accuracy because its errors are less correlated with the local models.

\FloatBarrier
\subsection{Semantic fidelity versus in-domain fit}
\label{sec:stancefix}

The submitted judge inherits the stance-handling error described in Section~\ref{sec:stance}. To isolate its effect, we reran rubric v3 with only the stance-dependent instructions and explanations corrected to the published payoff definition; all few-shot pairs and labels, model settings, and decoding parameters were otherwise held fixed.

The semantically corrected rubric performs \emph{worse} in-domain, scoring 0.622 versus 0.680 on the same 553 development pairs ($-5.8$ points; McNemar $p=0.001$). The effect is concentrated on the 127 pairs containing at least one bearish post, where accuracy falls from 0.772 to 0.575 ($-19.7$ points; $p=0.0002$); on the other 426 pairs, the change is only 1.6 points and lies within our development-noise range.

This inversion is more than a bug report. The original rule was semantically wrong but empirically useful because, in a predominantly bullish training regime, discounting bearish posts often improved the realized ranking. Correcting the payoff semantics removes that regime shortcut and forces the judge back toward the intended forecasting problem. Thus, validation accuracy alone would have selected the \emph{less faithful} rule. The result illustrates a general benchmark risk: when target outcomes are regime-dependent, an LLM rubric can become better calibrated to the local empirical distribution precisely by becoming less faithful to the task's intended semantics. Section~\ref{sec:postmortem} shows that this shortcut does not transfer to the official test set.
\FloatBarrier
\subsection{Human expert calibration}
\label{sec:human}

A native-speaker author, blind to model outputs, answered 38 of 40 labeled training pairs (two left blank), scoring $23/38=0.605$, better on small-gap than large-gap pairs; overrides of the three-track majority would have fixed 4 machine errors and introduced 8, so human effort went to error analysis, rubric construction, and lexicon development instead.

\section{Official Results and the Transfer Collapse}
\label{sec:postmortem}

On the 87 official FinArg-3 pairs, our three submitted runs score 0.5172 (majority vote), 0.5172 (attribute majority), and 0.5057 (local blend). The top NTCIR-19 run reaches 0.5402, the organizers' constant all-Post-1 baseline scores 0.5287, and the eleven submitted runs span 0.4598--0.5402~\cite{chen2026overview}. At $n=87$, the binomial standard error around chance is 5.4 points, placing every submitted run inside the approximate fair-coin 95\% interval $[0.395,0.605]$.

The striking fact is the \emph{change of regime}, not merely the low score. In development, our ensemble reaches 0.734 and unanimous three-track decisions are about 0.75 accurate; officially, the 26 unanimous pairs score exactly 0.500. Order consistency likewise ceases to discriminate reliably, and the individual tracks fall to 0.483--0.529. These signals would be expected to stay useful under added independent noise, yet they collapse together.

The ERAI predecessor, whose BERT-Chinese system reached 0.6207 on a blind pairwise MPP test~\cite{gon2022jetsons}, makes a universal impossibility explanation implausible, although the absence of pair-level identity between the editions rules out a controlled comparison. The NTCIR-19 result is therefore better framed as a \emph{transfer collapse}: signal exploitable in one development regime, and historically in a predecessor benchmark, does not reliably carry into the present official pairs.

Our own semantic error does not explain the collapse. Correcting the bearish rule changes 28 of 87 judge predictions but leaves official judge accuracy at 0.4828; the changed decisions split 14--14 between gains and losses. The same correction that loses 5.8 development points is therefore neutral on the official set, consistent with a regime-specific shortcut that ceases to pay off out of domain.

A second clue is outcome margin. Development accuracy rises with the labeled MPP gap: 0.620 in the lowest-gap tercile, 0.755 in the middle, and 0.816 in the highest. We do not claim that small gaps are intrinsically difficult for every predictor. Section~\ref{sec:theory} asks a narrower question: for our predictor, how strongly does reliability depend on outcome separation, and what can shift that separation?
\section{Real-Time Evaluation and Top-210 Ranking}
\label{sec:realtime}

The real-time evaluation (July 20--24) delivered ten fresh pairs per day with a 24-hour response window. It coincided with an unusually volatile episode, a record crash on July 17 followed by a record rebound and a second decline, so the batches reflect a pronounced crash-and-rebound regime.

We froze the offline system and deployed it through a deterministic daily pipeline that records all track probabilities and decision modes and falls back to the local blend whenever the LLM judge abstains or the API fails. Across the five days the judge's two presentation orders agreed on 7--10 of the 10 daily pairs, and human intervention was restricted to near-tie cases with an explicit structural justification and agreement of both team members; only two such overrides were exercised. Although the 24-hour window exposes one trading day of realized price movement, this never triggered an override, so the real-time submissions are effectively those of the frozen system.

\FloatBarrier
\subsection{Top-210 ranking via observed lower bounds}

The supplementary Top-210 task asks systems to rank the 210 posts with highest expected MPP from a 2{,}100-post July 2026 collection. Because partial price paths were observable for many posts before the ranking deadline, we augmented the frozen text score with an observed favorable-excursion lower bound: an LLM extracts the target ticker and any quoted entry price, after which daily market data measure the excursion observed since posting.\footnote{Price data here and in Section~\ref{sec:theory}: Yahoo Finance via \texttt{yfinance}, dividend-unadjusted daily bars, cached 25 July 2026 for the submitted ranking and re-retrieved 22 August (truncated to 24 July) for the post-hoc analysis. Entry proxy: the author-quoted price when within 15\% of the post-date close, else that close. Post-hoc excursions begin the next trading day; the submitted ranking included the posting day. Because these choices do not reproduce full-horizon MPP, Section~\ref{sec:theory} treats the outcome as a truncated-horizon MPP proxy.}

The submitted implementation used our then-held long-only interpretation. Under the correct stance-aware definition, 53 of the 210 submitted entries are bearish and were scored on the wrong side; we therefore treat the submitted ranking as an operational result, not a stance-correct estimate of MPP. Quoted prices were accepted only within 15\% of the post-date close and infeasible compounded returns were rejected; after these checks, 166 of 210 submitted entries were supported by observed price data; the remainder used the frozen model score.

After our experiments concluded, the organizers published a public dashboard for both real-time tracks. Table~\ref{tab:dashboard} reports the final standings for the evaluation window ending 24 August 2026.\footnote{Organizer dashboard: \url{https://haalab.github.io/sites/finarg3_realtime.html}.} The two tracks separate exactly along the decomposition of Section~\ref{sec:theory}: every team's pairwise accuracy over the 50 live pairs lies within binomial noise of chance (0.39--0.51, ours highest), whereas magnitude ranking separates teams clearly and our observed-excursion method leads at 10.58\% average portfolio MPP against the 6.18\% all-pool baseline.

\begin{table}[htbp]
\centering
\small
\caption{Final organizer dashboard standings for the two real-time tracks, with the evaluation window ending 24 August 2026. RT acc.\ is pairwise accuracy over the 50 live pairs, with ties receiving half credit; the Top-210 portfolio treats each selection as 210 equal slots.}
\label{tab:dashboard}
\centering
\small
\begin{adjustbox}{max width=\linewidth}
\begin{tabular}{lccc}
\toprule
Team & RT acc. & RT avg.\ MPP & Top-210 MPP \\
\midrule
BAIBAICHUCHU (ours) & \textbf{0.51} & \textbf{16.47\%} & \textbf{10.58\%} \\
MouMouRUC & 0.47 & 14.02\% & 8.23\% \\
UTaipei & 0.45 & 15.89\% & --- \\
SCUNLP & 0.45 & 14.33\% & 9.92\% \\
TMUNLP & 0.39 & 13.45\% & 5.40\% \\
\midrule
All-2{,}100 baseline & --- & --- & 6.18\% \\
\bottomrule
\end{tabular}
\end{adjustbox}
\end{table}

\section{Decomposing Profitability Ranking: Volatility, Horizon, and Pairwise Margin}
\label{sec:theory}

The transfer collapse suggests that MPP ranking is not pure text classification. Two outcome-side components can change while the language model is fixed: volatility plus observation horizon, and the \emph{margin} separating two outcomes.

To relate volatility to favorable-excursion magnitude, let the log price follow a driftless Gaussian random walk with daily volatility $\sigma$ over $T$ trading days. If $M_T$ is the running maximum of the log-price change, then
\[
\mathbb{E}[M_T]\approx\sigma\sqrt{\frac{2T}{\pi}}.
\]
By reflection symmetry, the running minimum has the same magnitude scaling, so the same $\sigma\sqrt{T}$ relationship applies to bullish and bearish favorable directions. We do not claim that prices are Gaussian random walks; the model is useful because it predicts a specific volatility--horizon scaling before any textual feature is introduced.
\FloatBarrier
\subsection{Volatility--horizon scaling in the July proxy outcomes}
The published MPP construction uses a prescribed entry and fixed evaluation horizon~\cite{chen2021rationales}; the organizer work cited for FinArg-3 sets $T=60$ trading days~\cite{chen2025opinion}. Our July reconstruction does not reproduce that full outcome. For each price-aligned post $i$, we instead compute a truncated-horizon MPP proxy (TH-MPP), $\mathrm{TH\text{-}MPP}_i(N_i)$, over the post-specific observable horizon $N_i$ ending 24 July 2026 (artifact field: \texttt{TFE}). It uses upward excursion for bullish and neutral posts and downward excursion for bearish posts; the neutral convention is specific to this proxy. Unlike official MPP, it also uses the proxy entry and dividend-unadjusted prices described in Section~\ref{sec:realtime}.

We first operationalize $\sigma_i$ as the per-ticker realized daily volatility of the July sample itself. Because that sample extends past the posting date, this is an \emph{ex-post structural test}: it asks only whether realized outcomes exhibit the predicted $\sigma_i\sqrt{N_i}$ scaling, and it motivates rather than establishes predictability. Across 502 posts with usable aligned windows, $\sigma_i\sqrt{N_i}$ correlates with $\mathrm{TH\text{-}MPP}_i$ at Spearman $\rho=0.399$ (bearish subset $\rho=0.554$), and a through-origin regression gives a slope of 0.78 against the driftless value $\sqrt{2/\pi}\approx0.80$. The outcome is truncated and returns are non-Gaussian, so the agreement should not be over-read; what it shows is that a theoretically specified combination of volatility and horizon organizes much of the proxy-outcome magnitude. The more important question is whether this relationship is measurable using pre-posting volatility and the post-specific observation horizon.

\FloatBarrier
\subsection{Pre-posting volatility is informative ex ante}
\label{sec:exante}
For each post we define $\sigma_{\mathrm{pre},20}$ as the sample standard deviation of the last 20 daily log returns of the adjusted close whose return end dates fall strictly before the posting date, so no post-date price enters; volatility is daily, not annualized, and $N_i$ is the same post-entry horizon as above. The corresponding horizon-scaled volatility score is $\sigma_{\mathrm{pre},20}\sqrt{N_i}$. Because posts sharing a ticker are dependent, all intervals are ticker-cluster bootstrap 95\% CIs (5{,}000 replicates). The 20-return lookback was fixed as primary and a 60-return lookback as robustness before any result was seen.

\begin{table}[htbp]
\centering
\small
\caption{Pre-posting volatility, horizon, and July $\mathrm{TH\text{-}MPP}$ (Spearman $\rho$ with ticker-cluster bootstrap 95\% CI). $n=502$ posts over 86 tickers; the 60-return-lookback row uses 496 posts over 83 tickers.}
\label{tab:exante}
\centering
\begin{adjustbox}{max width=\linewidth}
\begin{tabular}{lcc}
\toprule
Predictor & $\rho$ & 95\% CI \\
\midrule
$\sqrt{N_i}$ & 0.202 & $[0.090,\,0.345]$ \\
$\sigma_{\mathrm{pre},20}$ & 0.209 & $[0.015,\,0.348]$ \\
$\sigma_{\mathrm{pre},20}\sqrt{N_i}$ (primary) & \textbf{0.320} & $[0.133,\,0.466]$ \\
$\sigma_{\mathrm{pre},60}\sqrt{N_i}$ (robustness) & 0.385 & $[0.224,\,0.491]$ \\
\midrule
partial: $\sigma_{\mathrm{pre},20}$ given $\log N_i$, stance & 0.281 & $[0.070,\,0.448]$ \\
\bottomrule
\end{tabular}
\end{adjustbox}
\end{table}

Raw 20-return-lookback volatility correlates with later $\mathrm{TH\text{-}MPP}$ at $\rho=0.209$; the horizon-scaled 20- and 60-return-lookback scores reach 0.320 and 0.385. The product has a larger point association than either component, but overlapping intervals do not establish superiority. The partial rank correlation of $\sigma_{\mathrm{pre},20}$ with $\mathrm{TH\text{-}MPP}$ given $\log N_i$ and stance is 0.281, so horizon alone does not explain the volatility association. The primary product is associated in both directional subsets, descriptively more strongly for bearish posts.

For pairwise ordering, $\sigma_{\mathrm{pre},20}\sqrt{N_i}$ gives Kendall $\tau_b=0.229$; among 124{,}996 unique pairs with unequal $\mathrm{TH\text{-}MPP}$, the higher score corresponds to the larger proxy outcome in 61.4\% (CI $[0.552,\,0.667]$). This is July proxy-pair ranking, not official FinArg-3 accuracy. On the same 502 posts, the ex-post association is $\rho=0.399$, versus 0.320 for the primary pre-posting score (pre-posting minus ex-post $=-0.078$, CI $[-0.189,\,-0.003]$). Removing post-publication information weakens but does not eliminate the association; differing volatility lookbacks mean the gap is not purely a leakage effect (the 60-return-lookback value nearly matches the ex-post one).
\FloatBarrier
\subsection{Pairwise reliability depends on the realized margin}
For a development pair, let $\Delta=\mathrm{MPP}_A-\mathrm{MPP}_B$ and $Y=\operatorname{sign}(\Delta)$. We model our predictor's latent gap estimate as
\[
\widehat{\Delta}=\Delta+\epsilon,\qquad \epsilon\sim\mathcal{N}(0,s^2),
\]
so that
\[
P(\widehat{Y}=Y\mid\Delta)=\Phi\!\left(\frac{|\Delta|}{s}\right).
\]
This is a predictor-specific descriptive model, not an information-theoretic ceiling; a better-informed predictor corresponds to a smaller effective $s$.

\begin{figure}[htbp]
    \centering
    \includegraphics[width=\linewidth]{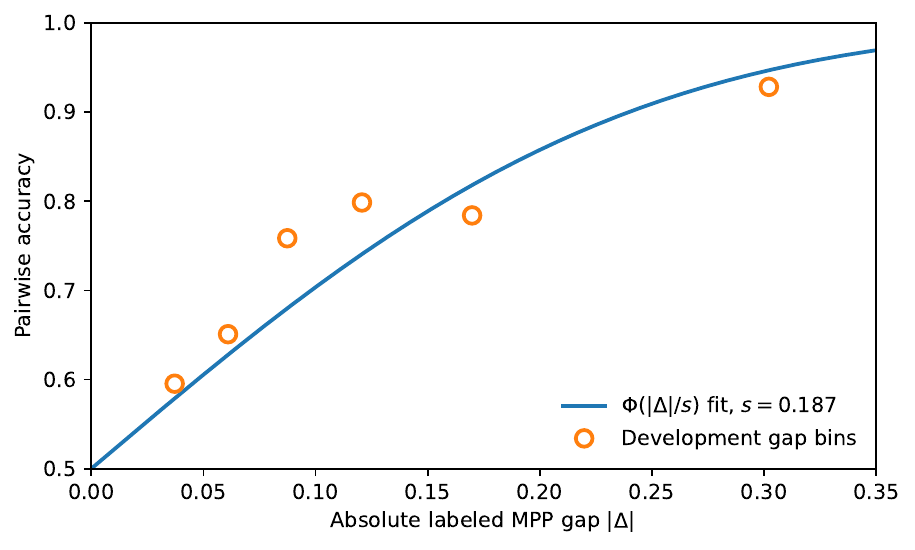}
    \caption{Development pairwise accuracy increases with the absolute labeled MPP gap. Circles are six equal-count gap bins from the ensemble predictions; the curve is the one-parameter fit $\Phi(|\Delta|/s)$ with $s=0.187$. The figure describes our predictor on the development distribution and makes no claim about the unavailable official gap distribution.}
    \Description{A curve of pairwise accuracy versus absolute labeled MPP gap. Six empirical development bins rise from about 0.60 accuracy at small gaps to about 0.93 at the largest gap, with a fitted Gaussian reliability curve.}
    \label{fig:gap-reliability}
\end{figure}

Maximum-likelihood fitting gives $s=0.187$. The six empirical bins follow the resulting curve closely: accuracy is about 0.60 in the narrowest-gap bin and rises to about 0.93 in the widest. Figure~\ref{fig:gap-reliability} therefore supplies a quantitative version of a simple fact: for this predictor, the same amount of prediction error is much more likely to flip the ordering when the two realized MPP values are close.

The separate July reconstruction asks whether shared-asset comparison can shift the \emph{distribution} of outcome margins. Among 502 usable posts, 5{,}531 same-ticker pairs have a median absolute $\mathrm{TH\text{-}MPP}$ gap of 4.1 percentage points versus 5.2 across 120{,}220 cross-ticker pairs; 35.4\% versus 21.4\% have gaps below two points. Same-ticker, same-stance pairs are narrowest at 2.7 points median, compared with 5.0 for opposite-stance same-ticker pairs. We deliberately do not substitute these proxy gaps into the development reliability curve because the outcome construction differs. The defensible conclusion is distributional: pair construction can concentrate a benchmark in a region where a fixed predictor is less reliable. The corresponding pre-posting volatility--horizon analysis cannot be replicated on the official pairs because the released posts carry no posting timestamps to which a strictly pre-posting volatility window could be aligned; that association is therefore evaluated only in the timestamped July collection.
\FloatBarrier
\subsection{Admissible text signals}
Under a semi-strong market-efficiency perspective~\cite{fama1970efficient}, persistent predictive value from public text should be limited once public information is priced. Compatible channels include possible private-information leakage (the ``dream-post'' idiom, $r=+0.40$ in isolation); instrument identification, whose contribution is not attributed here to volatility; and contemporaneous state revelation. Table~\ref{tab:attrs} aligns with these channels, but the associations are exploratory, not causal.

\FloatBarrier
\subsection{How much signal do two coarse attributes capture?}
\label{sec:residual}
A two-feature baseline using target asset class and an LLM-assessed volatility tier---not measured volatility---gives 0.591 development accuracy, versus 0.682 for the full text model. Relative to chance, these text-derived features recover approximately half of the full model's improvement,
\[
\frac{0.591-0.5}{0.682-0.5}\approx0.50.
\]
The remaining margin indicates in-domain information beyond these proxies, without identifying which textual components provide it.

\begin{sloppypar}
The July collection permits a cross-period comparison with the pre-posting volatility--horizon score of Section~\ref{sec:exante}. The frozen 2019--2020 text score is nearly uncorrelated with $\mathrm{TH\text{-}MPP}$ ($\rho=0.055$, CI $[-0.060,\,0.157]$), versus $\rho=0.320$ for $\sigma_{\mathrm{pre},20}\sqrt{N_i}$. Given this non-textual score and stance, text remains at $\rho=0.079$ (CI $[-0.047,\,0.181]$); given text and stance, the volatility--horizon score is $\rho=0.368$ (CI $[0.190,\,0.518]$). Their correlation is $-0.061$, so collinearity does not explain the partials. Pairwise, the volatility--horizon score reaches 61.4\% (CI $[0.552,\,0.667]$), text 51.9\% ($[0.479,\,0.554]$), and their no-fit equal-rank sum 59.3\% ($[0.546,\,0.631]$). An in-sample rank regression reaches 0.654 with both versus 0.652 with the non-textual score. Thus transferred text adds little beyond horizon-scaled pre-posting volatility in this July sample, where the relationship remains visible.
\end{sloppypar}

\section{Discussion}
\label{sec:discussion}
Profitability ranking is composite. The \emph{textual} component expresses direction, state, instrument, and possibly unpriced information. The \emph{market-based} component combines pre-posting volatility with the post-specific July observation horizon. \emph{Comparison geometry} determines whether a benchmark asks the predictor to resolve narrow or wide outcome margins.

This view reconciles three observations. A wrong bearish rule wins in development by encoding a local regime prior, so semantic fidelity and validation fit differ. In-domain, text and semantic heuristics are predictive (ensemble 0.734; the 0.591 two-feature baseline recovers about half of the 0.682 full text model's margin over chance); across periods, transferred text is near zero while the pre-posting volatility--horizon score remains associated within July. The gap--accuracy curve also shows one predictor varying sharply across outcome margins. Thus signals can change across regimes: textual heuristics are local, volatility and horizon correlate with July outcomes, and reliability depends on outcome separation. These relationships do not explain the full official collapse because official pair composition and MPP gaps are unavailable.

The ERAI comparison sharpens rather than weakens this conclusion. Above-chance performance in the predecessor task means the NTCIR-19 leaderboard is not evidence that MPP ranking is inherently impossible; rather, such benchmarks are sensitive to the joint distribution of text, assets, regimes, and comparison pairs. Cross-edition reporting should therefore include label construction, horizon and entry conventions, outcome-margin distributions, uncertainty against simple baselines, and non-textual market baselines.

Future work should test whether the pre-posting volatility association persists on full-horizon MPP labels and across market regimes, and should add other pre-posting market covariates such as turnover and drawdown.

\FloatBarrier
\subsection*{Limitations}
The submitted LLM judge used an incorrect long-only treatment of bearish posts, and 53 of the 210 submitted Top-210 entries were scored on the wrong side; a system built stance-aware from the outset remains untested. Development contains only 91 unique labeled posts and was iteratively tuned, so 0.734 is tuned development performance. July $\mathrm{TH\text{-}MPP}$ is not official 60-day MPP: it uses a post-specific truncated horizon, proxy entry, dividend-unadjusted OHLC, and an upward convention for neutral posts (an ex-dividend date falls within 29 of the 502 windows), whereas the pre-posting volatility uses adjusted closes and strictly pre-posting returns. The collection covers a single short, unusually volatile regime, and the text--volatility--horizon comparison is cross-period rather than general. Finally, official pair construction and MPP gaps remain unavailable, so the volatility--horizon and margin relationships cannot be cleanly separated from other forms of distribution shift.
\section{Conclusions}
The most important result of our FinArg-3 participation is not the 0.517 official score by itself, but the contrast between a 0.734 tuned development ensemble, unanimous official predictions at exactly 0.500, a near-chance NTCIR-19 leaderboard, and an earlier ERAI result substantially above chance. This pattern turns the task from a simple system comparison into a question about when profitability-ranking signal transfers.

Our post-hoc analyses identify three relationships. A semantic audit shows that an incorrect bearish payoff rule can be more predictive in-domain because it encodes a regime shortcut. A running-extremum model predicts $\sigma\sqrt{T}$ scaling; $\sigma_{\mathrm{pre},20}\sqrt{N_i}$ is associated with later July $\mathrm{TH\text{-}MPP}$ and pair order, unlike transferred text. A predictor-specific reliability model shows that pairwise accuracy depends strongly on the realized MPP margin.

Profitability ranking is not simply argument understanding. In our data, strong in-domain signals can fail to transfer, while a pre-posting horizon-scaled volatility score remains associated with later outcome magnitude and pairwise reliability depends on outcome separation. MPP is not inherently unpredictable; evaluation should jointly consider text, regime, historical volatility, horizon, and pair construction.
\bibliographystyle{plainnat}
\bibliography{refs}

\end{document}